\documentclass[10pt,journal,compsoc]{IEEEtran}
\usepackage{graphicx}
\usepackage{algorithm}
\usepackage{algpseudocode}
\usepackage{blindtext}
\usepackage{float}
\usepackage{amsmath}
\usepackage{mathtools}
\usepackage{eqnarray,amsmath,amssymb}
\ifCLASSOPTIONcompsoc
  \usepackage[nocompress]{cite}
\else
  \usepackage{cite}
\fi
\ifCLASSINFOpdf
\else
\fi

\begin{document}
%
\title{A Fully Controllable Agent in the Path Planning using Goal-Conditioned Reinforcement Learning}
%
%
%
%

\author{GyeongTaek~Lee,~\IEEEmembership{Member,~IEEE,}
\IEEEcompsocitemizethanks{\IEEEcompsocthanksitem GyeongTaek Lee is with Department of Industrial and Systems Engineering, Rutgers University, The State University of New Jersey, 96 Frelinghuysen Road, Piscataway, 08854, NJ, USA\protect\\
E-mail: gyeongtaek.lee@rutgers.edu
}
\thanks{Manuscript received May 19, 2022; revised August 26, 2022.}}

%
%

\markboth{Journal of \LaTeX\ Class Files,~Vol.~14, No.~8, August~2015}%
{Shell \MakeLowercase{\textit{et al.}}: Bare Demo of IEEEtran.cls for Computer Society Journals}
%



\IEEEtitleabstractindextext{%
\begin{abstract}
The aim of path planning is to reach the goal from starting point by searching for an agent’s route. In the path planning, the routes may vary depending on the number of variables such that it is important for the agent to reach various goals. Numerous studies, however, have dealt with a single goal that is predefined by the user. In the present study, I propose a novel reinforcement learning framework for a fully controllable agent in the path planning. To do this, I propose a bi-directional memory editing to obtain various bi-directional trajectories of the agent, in which the agent’s behavior and sub-goals are trained on the goal-conditioned RL. As for the agent’s to move in various directions, I utilize the sub-goals dedicated network, separated from a policy network. Lastly, I present the reward shaping to shorten the number of steps for the agent to reach the goal. In the experimental result, the agent was able to reach the various goals that have never been visited by the agent in the training. We confirmed that the agent could perform difficult missions such as a round trip and the agent used the shorter route with the reward shaping.


\end{abstract}

\begin{IEEEkeywords}
Controllable agent, path planning, goal-conditioned reinforcement learning, bi-directional memory editing.
\end{IEEEkeywords}}

\maketitle

\IEEEdisplaynontitleabstractindextext

%
\IEEEpeerreviewmaketitle

\IEEEraisesectionheading{\section{Introduction}\label{sec:introduction}}
\IEEEPARstart{P}{ath} planning is a method to find an optimal route from the starting point to the target point. It has been widely used in various fields such as robotics \cite{yu2020path,kumar2020comparison,wohlke2021hierarchies}, drone \cite{jeauneau2018path,qie2019joint,wang2020multi,wang2020mobile,hayat2020multi,bayerlein2021multi}, military service \cite{lee2020autonomous,hu2022autonomous}, and self-driving car \cite{wang2019obstacle,tammvee2021human}. Recently, reinforcement learning (RL) has been mainly studied for the path planning \cite{lee2020autonomous,wang2020mobile,yao2020path,qiao2020hierarchical,bayerlein2021multi,lin2021collision,cao2021confidence,wohlke2021hierarchies,hu2022autonomous}. To get an optimal solution, it is essential to give enough reward for an agent to reach the goal and to set up a specific environment. Several studies on learning the RL model have proposed to make an agent be robust in a complicated or an unknown environment for the path planning \cite{li2019deep,yang2020efficient,hu2020voronoi}. However, existing studies have defined one single goal before the learning. That is, the agent’s ability to search the path when completed learning can be limited. To make the agent reach a number of goals in a dynamic environment, learning a controllable agent is needed. One of the recent approaches for controlling the agent has a limitation in that the agent can only learn the behavior from trajectories that have been directly experienced \cite{lee2022learning}. Therefore, the agent can only be under control in the area visited by the agent. In this paper, I focus on learning a fully controllable agent in the path planning using a goal-conditioned RL. Especially, I apply, to the goal-conditioned RL, a bi-directional memory editing and a sub-goals dedicated network to improve the ability to search the path of the agent. 

\begin{figure}[t]
\centering
\includegraphics[scale=0.40]{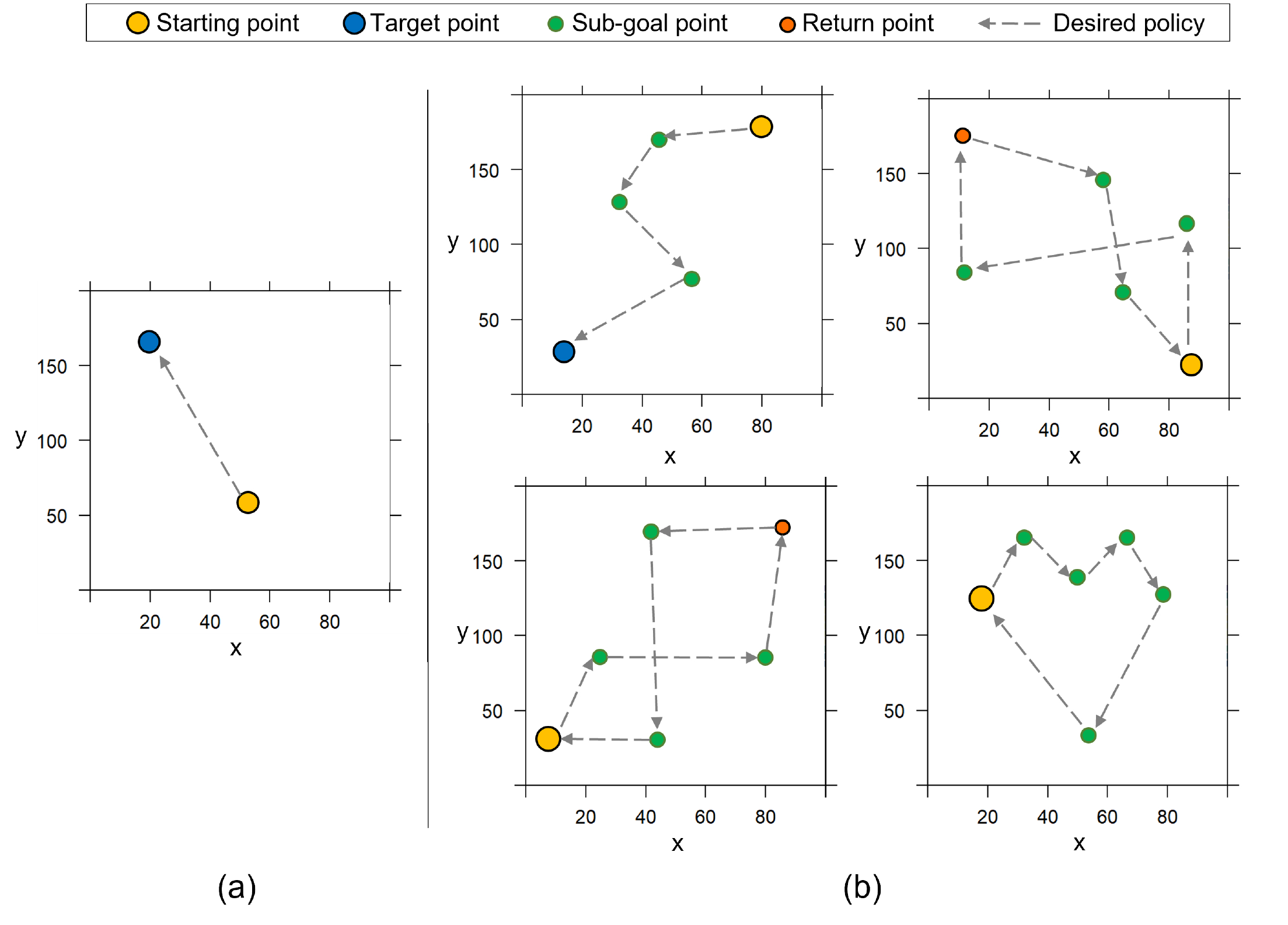}
\caption{Examples of training and test environment in this study. The experiment was set as a simple version of the two-dimensional grid environment. The starting point and the target point were fixed at the location in the training environment. The policy of the agent is to search the path from the starting point to the target point. (a), The training environment was set for the agent to easily get an optimal route. (b), The test environment was set for the agent to get an optimal route in a difficult way. Several sub-goals including the final goal were given to the agent.  }
\label{fig1}
\vskip -0.2in
\end{figure}

In the goal-conditioned RL, the agent learns the sub-goals, part of the trajectory of the agent \cite{lee2020weakly,okudo2021subgoal,zhang2021world,chane2021goal,yang2022rethinking}. By learning the sub-goals, eventually, the agent can reach the final goal. This method showed good performance mainly in robotics. However, previous studies have focused on reaching a complicated single goal. In addition, the multi-goal RL model, in which the agent can perform many goals, has been proposed \cite{bai2019guided,zhao2019maximum}. However, the multi-goals should also be defined before the learning. Unlike these studies, in this present study, I propose an RL framework in which the agent can perform a number of sub-goals in various scenarios. An important point here is that the sub-goals are not defined in advance. In other words, the agent completed the learning reaches the goals, which have never been visited by the agent in the learning.

Fig \ref{fig1} shows the examples of the training and the test environments of this study. The training environment has been simply set for the agent to reach the final goal. But the test environments have been set in a difficult way. The agent should go through the sub-goals and reach the target point, even if the starting point and the target point are set in reverse. The difficult missions were also given to the agent such as a round trip. In the test environments, an important thing is that the agent has to reach the goal never been visited in the learning. To perform the goals, the agent must be fully controlled by the sub-goals that are customized by the user.

 Meanwhile, memory editing occurs in our daily lives. \cite{bernstein2009tell,schacter2011memory,fernandez2015benefits,phelps2019memory}. The memory editing helps us to get away from mental illnesses such as trauma \cite{phelps2019memory}. We can also get precise information by editing untidy memories \cite{fernandez2015benefits}. A recent study used the concept of the memory editing on goal-conditioned RL for the agent to reach sub-goals so that the user can control the agent \cite{lee2022learning}. However, the agent cannot move the sub-goals in difficult environments. This is because the memory of the RL model is edited only based on one direction of the paths that are moved by the agent. The first purpose of the RL is to achieve the final goal. Despite the sub-goals, the agent can ignore the sub-goals if the sub-goals get in the way of the final goal. That is, the agent does not need to go round and round to reach the final goal. In this study, I developed the concept of the memory editing for the fully controllable agent in the path planning. 
 
\begin{figure}[t]
\centering
\includegraphics[scale=0.6]{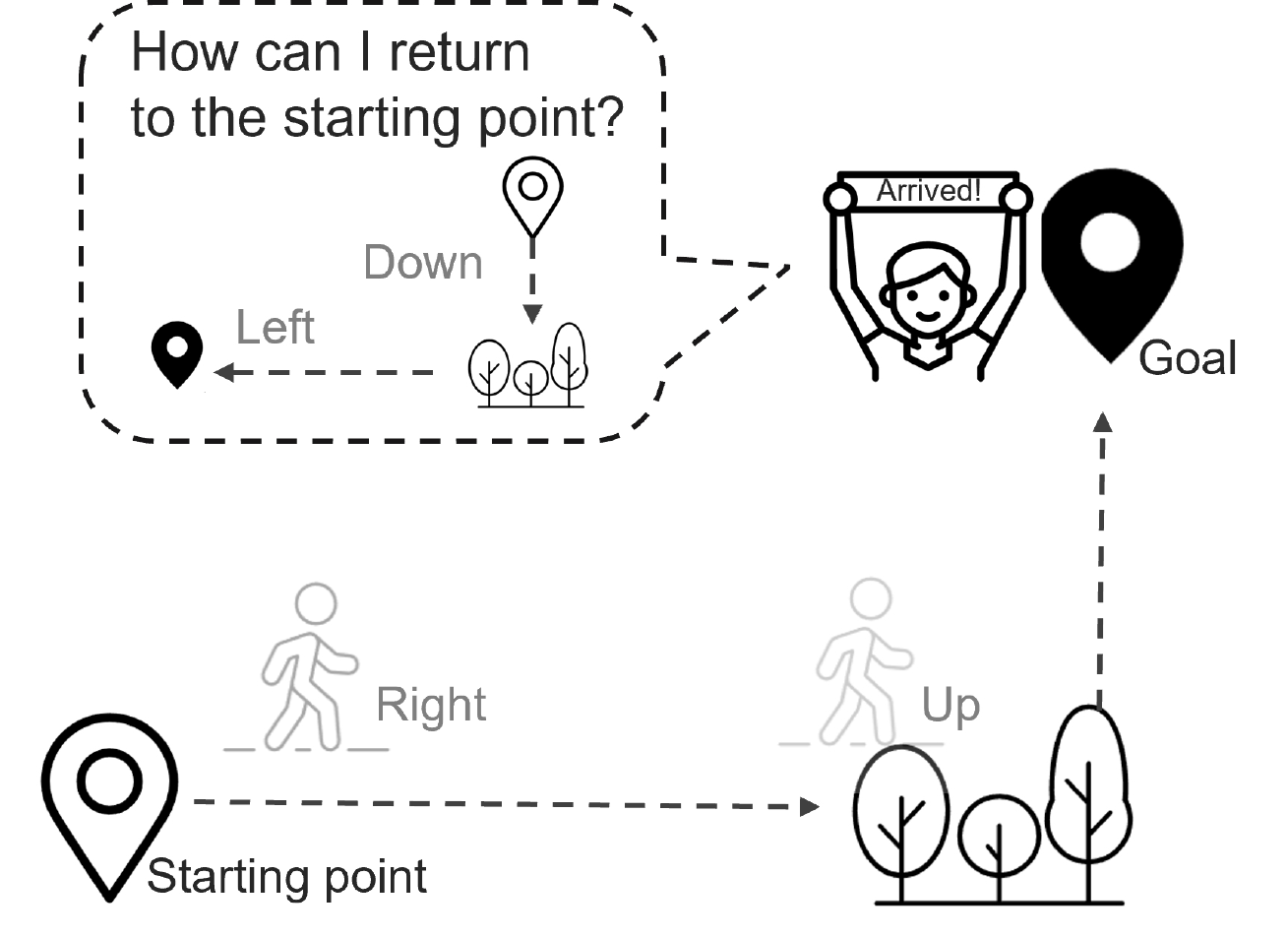}
\caption{ Illustration of the concept of the bi-directional memory editing. After we arrive at the destination, if we recall our memory, we can find out the route for returning to the starting point. Further, we can also obtain knowledge about which action to perform to reach the starting point.    }
\label{fig0}
\end{figure}

Let’s assume that we walk to the destination (See Fig \ref{fig0}). From the route that we walked, we can learn how we reach the destination. We can remember intermediate stops in the route and know that the total path is the overall set, which consists of the intermediate stops. Further, if we recall our memories backward, we can also find the route back to the starting point. For example, in the Fig \ref{fig0}, when we want to return to the starting point, we know which action we have to perform. However, it is difficult to find an inverse action in a dynamic RL environment. Thus, an inverse module to predict the inverse action is necessary to obtain the exact knowledge to come back to the starting point. 

Based on the processes, in which we recall our memories and obtain knowledge, I utilize a bi-directional memory reminiscence and editing (bi-directional memory editing) to obtain various trajectories. Using these trajectories, the agent can learn various behaviors. Furthermore, I use the dedicated network for learning the sub-goals to improve learning efficiency. Finally, I present a reward shaping for the shorter path of the agent. Using these techniques, the agent can achieve various sub-tasks as well as the final goal in the path planning environment. The fully controllable agent can be useful in the environments where we have to consider a number of variables and we have to assume various scenarios. The main contributions of this article are as follows:

\begin{itemize}
\item Using the bi-directional memory editing, we can obtain various trajectories, and can learn the agent to perform various tasks, based on the trajectories. The agent can be fully controlled so that the agent can reach any point in the path planning environment. \\
\item I employ the sub-goals dedicated network to improve the efficiency of learning the sub-goals. To distinguish the network from the policy network, the agent can focus on performing the various sub-goals.  \\
\item I propose the reward shaping for the shorter path of the agent. In the path planning, it is important for the agent not only to reach the final goal but also to reach it within a limited time. By applying the reward shaping in the bi-directional memory editing, the agent can reach the final goal within a shorter time. \\
\item To the best of our knowledge, this study is the first RL methodology for the path planning in that the agent is fully under control. Therefore, the agent achieves the user-defined sub-goals such as a round trip in a difficult test environment. Moreover, the agent can move to the point that has never been reached by the agent in the training. By using this methodology, we can suppose and conduct various scenarios in the path planning. 
\end{itemize}

The rest of the paper proceeds as follows. Section 2 presents the background of this study. In Section 3, learning a fully controllable agent is proposed. To do this, I introduce the bi-directional memory editing and the sub-goals dedicated network. Furthermore, I propose the reward shaping for the shorter path of the agent. Section 4 provides the results of the experiment. I confirmed whether the agent can successfully perform the sub-goals in various scenarios. Section 5 concludes this paper and calls for future research.

\section{Background}
\subsection{Path planning}

To get an optimal solution to reach the target point from the starting point, traditionally, optimization methods have been used in the path planning. Several studies using an A* \cite{yan2018path,chen2020improved}, a genetic algorithm \cite{zhou2020trajectory}, and a particle swarm optimization \cite{huang2018uav,chen2021three, wang2022improved} have been proposed. Combining two optimization methods also has been studied \cite{jamshidi2020analysis}. 

\begin{figure*}[t]
\centering
\includegraphics[width=\textwidth]{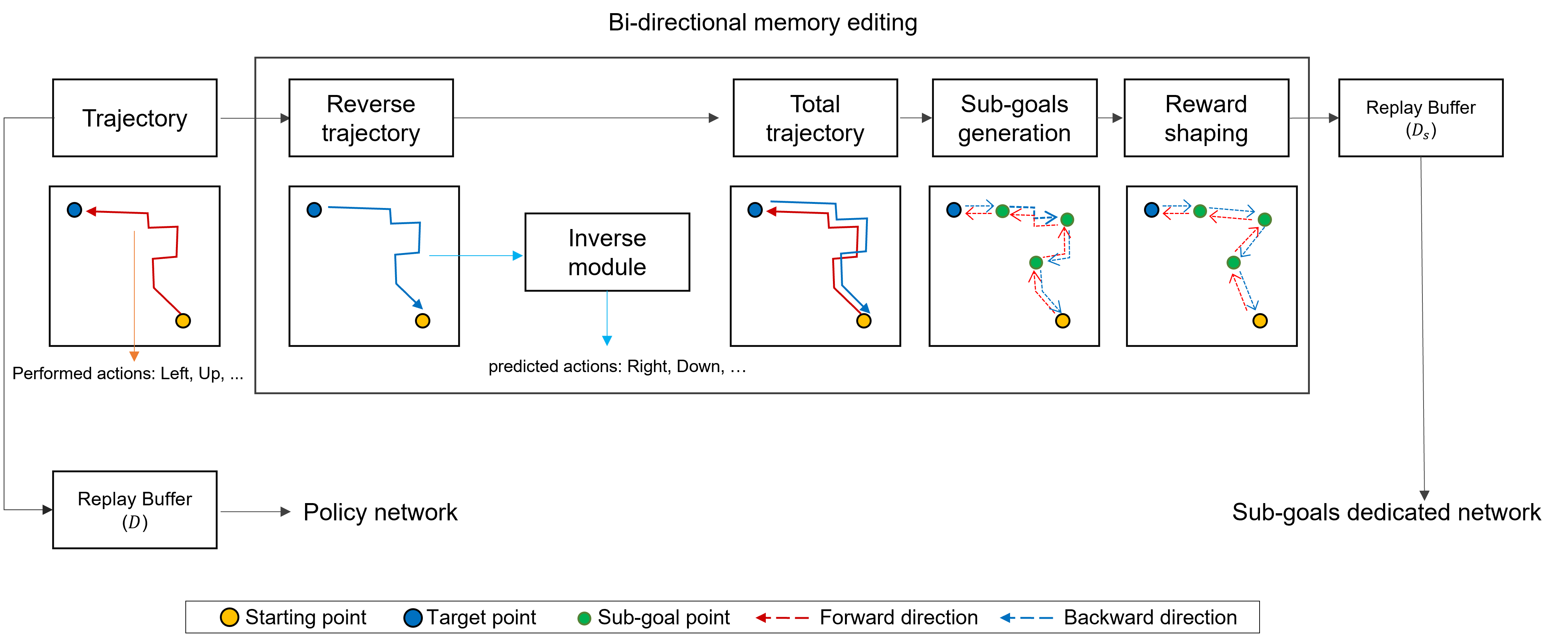}
\caption{Illustration of the proposed method. The red line indicates the route visited by the agent. The blue line indicates the inverse direction of the route. The inverse module predicts the action when $s_t$ and $s_{t+1}$ are given, and the predicted action is used to collect the reverse trajectories. In the bi-directional memory editing, various sub-goals are generated and stored in the separated replay memory $(D_s)$. The transition samples from $D_s$ are trained by the sub-goals dedicated network.}
\label{fig2}
\end{figure*}

Recently, with the development of deep learning, studies on the path planning using the RL have mainly been proposed \cite{lei2018dynamic,lee2020autonomous,liu2021novel,wang2020mobile,yao2020path,yan2020towards,wang2020multi,qiao2020hierarchical,bayerlein2021multi,lin2021collision,cao2021confidence,wohlke2021hierarchies,hu2022autonomous}. They have supposed the specific scenario and set an environment to apply the agent in the path planning. Especially, they have applied their study to robotics \cite{wang2020mobile,wohlke2021hierarchies,lin2021collision}, 
drone \cite{jeauneau2018path, qie2019joint ,yan2020towards, wang2020multi,hayat2020multi,bayerlein2021multi }, and ship \cite{chen2019knowledge, guo2020autonomous}. Also, they have focused on the one single goal of the agent to reach the target point, avoiding obstacles. After learning is completed, the agent could reach the goal; however, the agent cannot be under control in the previous studies. That is, the agent can only perform the predefined goal. 

In addition, learning user-defined sub-goals has been proposed \cite{lee2022learning}. However, the agent was partially under control. The agent could not perform the round trip and only moved the area that was visited by the agent. In this study, I focus on learning the fully controllable agent so that the agent can perform various trips, as shown in Fig \ref{fig1}.b.

\subsection{Goal-conditioned RL}

Hindsight experience replay (HER) used the sub-goals and the pseudo rewards to make the RL model converge to the final policy \cite{andrychowicz2017hindsight}. The reason why learning the sub-goals improves the performance of the RL is that the sub-goals are located on the route to reaching the final goal. Several studies developed the HER for an exploration of the agent \cite{nguyen2019hindsight, fang2019curriculum,lai2020hindsight }. 

The goal-conditional RL models excel to reach the goal through the intermediate sub-goals and show excellent performance on the robotics problem \cite{nasiriany2019planning,zhao2019maximum,ghosh2019learning,nair2018visual,eysenbach2019search,bai2019guided,eysenbach2020c,lee2020weakly,okudo2021subgoal,zhang2021world,chane2021goal,yang2022rethinking}. They have focused on searching for meaningful sub-goals and improving the performance of the main policy network (high-level policy network). To reach the desired policy, it is important to induce the agent to reach the landmark of the sub-goals \cite{nair2018visual,nasiriany2019planning,bai2019guided,zhang2021world,kim2021landmark} and to improve sample efficiency \cite{nachum2018data,eysenbach2019search,gurtler2021hierarchical}. The existing studies with learning the sub-goals are similar to this study in terms of generating the sub-goals and relabeling the rewards.


However, in this study, the sample efficiency and searching for the valid sub-goals are not essential. By performing the bi-directional memory editing, we can get trajectories two times more than when bi-directional memory editing is not performed. Then, enough sub-goals can be collected for learning the sub-goals. Also, it does not matter the time when the sub-goals dedicated network is trained, which is separated from the original policy network. Whether the sub-goals are trained after learning the policy network or during learning the policy network, if sufficient trajectories are gathered, the agent can learn various behaviors and sub-goals. The main purpose of this study is to make the agent under control for performing various tasks which are not defined in the training environment. After learning is completed, the agent can achieve various user-defined sub-goals as well as the final goal.  

I consider a discounted, finite-horizon, goal-conditioned Markov decision process (MDP) defined as a tuple ($\mathcal{S}, \mathcal{G}, \mathcal{A}, p,R, \gamma, H$), where $\mathcal{S}$ is the set of state, $\mathcal{G}$ is the set of goals, $\mathcal{A}$ is set of actions, $p(s_{t+1}|s_{t},a_{t})$ is a dynamics function, $R$ is the reward function, $\gamma \in [0,1)$ is the discounted factor, and $H$ is the horizon. In the goal-conditioned RL, the agent learns to maximize the expected discounted cumulative reward $\mathbb{E}[ \sum_{t=1}^\infty R(s_t,g,t)]$, where $g$ is the goal. The objective is to obtain a policy $\pi(a_t|s_t,g,t)$.

\section{Proposed method}
Fig \ref{fig2} shows the illustration of a summary of the proposed method. For the fully controllable agent in the path planning, I introduce three simple techniques. First, I propose a bi-directional memory editing to generate various behaviors and sub-goals of the agent. Here, to secure reverse trajectories, an inverse module to predict actions is used. Second, to improve the efficiency of learning, I utilize the sub-goals dedicated network separated from the policy network. Finally, I present a reward shaping for the shorter path of the agent.

\subsection{Bi-directional memory editing}
The memory editing is performed to generate sub-goals of the agent. The sub-goals are generated from the trajectories of the agent, and additional rewards are given to the agent. As the agent begins to recognize the sub-goals, the agent can achieve the sub-goals and greedily reach the final goal in the previous studies.

In the path planning, it is important to allow the agent to visit a wider area such that the agent can visit various locations and learn the optimal route to reach the goal. Thus, if we can get various trajectories more than the trajectories that are actually visited by the agent, it can bring greater benefit to learning the agent’s ability for searching the path. In addition to a forward route, the reverse route from the goal to the starting point can be a useful ingredient for the agent to learn various behaviors and sub-goals. To do this, I employ the reverse trajectory to generate various sub-goals and to learn the robust RL model by performing the bi-directional memory editing. 

First, a forward memory editing is performed, which is described in line 21-24 in Algorithm 1. The sub-goals ($g$) are generated from the forward route, and the state of the sub-goals ($s_{t+1} \| g$) are made. After that, a backward memory editing is performed, which is described in line 25-29 in Algorithm 1. Reverse transition is $\{(s_{t+1},a_{t+1}',r_t,s_t)\}$ whereas original transition is $\{(s_t,a_t,r_t,s_{t+1})\}$. Here, it is difficult to obtain $a_{t+1}’$. The reason is that it is not simple to find an action to derive the $s_t$, when $s_{t+1}$ is given. I propose to use the inverse module to obtain $\hat{a}_{t+1}$ by predicting $a_{t+1}’$ when $s_t$ and $s_{t+1}$ are given. Like the forward memory editing, the sub-goals ($g$) and the state of the sub-goals ($s_{t+1} \| g$) are generated. Finally, the edited memories are stored in the replay memory for the sub-goals ($\mathcal{D_s}$).

Using the bi-directional memory editing, we can obtain two routes from the one single route moved by the agent. Beyond being two times the trajectories, it means that the agent can learn various relationships between the actions and the sub-goals. In the path planning, the agent can almost only learn the route that is visited by the agent. For instance, as shown in Fig \ref{fig1}.(a), the agent can almost learn the leftward and upward directions because of the location of the goal. However, using the bi-directional memory editing, the agent can learn all directions in various locations. Therefore, by learning a number of sub-goals and behaviors, the agent can reach the goals that were never visited in the training.

\begin{algorithm}[t]
\begin{algorithmic}[1]
\caption{Learning the sub-goals using bi-directional memory editing }\label{alg:algorithm1}

\State Initialize policy network parameters $\theta_{p}$
\State Initialize sub-goals dedicated network parameters $\theta_{g}$
\State Initialize inverse module parameters $\theta_{iv}$
\State Initialize replay buffer for original goal $\mathcal{D} \leftarrow \emptyset$
\State Initialize replay buffer for sub-goals $\mathcal{D_s} \leftarrow \emptyset$
\Procedure{Learning the sub-goals}{}
\For{episode = 1, M}
\State  \verb|\\| Simulation stage.
\For{each step}
\State Execute an action $s_t,a_t,e_t,s_{t+1} \approx \pi_{\theta}(a_t \mid s_t)$
\State Store transition $\mathcal{E}\leftarrow \mathcal{E} \cup \{(s_t,a_t,r_t)\}$
\State Optimize inverse module $\theta_{iv}$ 
\EndFor

\If{ $s_{t+1}$ is terminal}
\State Compute returns $R_t= \Sigma^\infty_{k}\gamma^{k-t}{r}_k$ in $\mathcal{E}$
\State $\mathcal{D}\leftarrow \mathcal{D}\cup \{(s_t,a_t,R_t)\}$
\State Clear episode buffer $\mathcal{E} \leftarrow \emptyset$
\EndIf
\State

\State   \verb|\\|Bi-directional memory editing
\State Generate sub-goals $g$ 
\State Generate state of sub-goals $s_{t} \| g$
 \algorithmiccomment{$\|$ denotes concatenation}
\State Set additional rewards $r_t'$ 
\State $\mathcal{D_s}\leftarrow \mathcal{D_s}\cup \{s_t \| g,a_t,r_t')\}$
\State Predict the inverse action $\hat{a_{t+1}} \leftarrow \theta_{iv}(s_{t+1},s_t)$ 
\State Generate reverse transition $\{(s_{t+1},\hat{a}_{t+1},r_t,s_{t})\}$
\State Generate sub-goals $g$ 
\State Generate state of sub-goals $s_{t+1} \| g$
\State Set additional rewards $r_t''$ 
\State $s_t \leftarrow s_{t+1}$,$a_t  \leftarrow \hat{a}_{t+1}$,$r_t  \leftarrow r_t''$
\State $\mathcal{D_s}\leftarrow \mathcal{D_s}\cup \{s_t \| g,a_t,r_t')\}$

\State
\State   \verb|\\|Learning stage.
\For{k= 1, N}
\State  Sample a minibatch $\{(s,a,R)\}$ from $\mathcal{D}$
\State \algorithmiccomment{Optimize policy network  $\theta_{p}$ }
\EndFor
\For{k= 1, P}
\State  Sample a minibatch $\{(s \| g,a,r')\}$ from $\mathcal{D_s}$
\State \algorithmiccomment{Optimize sub-goals dedicated network $\theta_{g}$ }

\EndFor
\EndFor
\EndProcedure

\end{algorithmic}
\label{algo1}
\end{algorithm}

\subsection{The sub-goals dedicated network}
Using the bi-directional memory editing, the agent can learn various behaviors and sub-goals. However, the agent at the middle of the route can be confused about where the agent has to go. Because the agent is forced to learn how to go in both directions at one point due to the bi-directionally edited memories. For example, in the Fig \ref{fig0}, if the agent is located near the tree, the agent learns to move both upwards and leftwards. If the agent is trained using the policy network, the agent can be confused about where the agent has to go if the agent is located near the tree. Because the purpose of the policy network is only to train the agent to reach the final goal. Actually, the agent has been hovering around the middle point of the environment in the experiment. Therefore, I employ the network for the sub-goals separately from the network for the final goal. 

The sub-goals dedicated network only learns the sub-goals, and the original policy network only learns the final goal. To do this, I also employ the replay memory for the sub-goals ($\mathcal{D_s}$) in addition to replay memory for the final goal ($\mathcal{D}$). Moreover, using the sub-goals dedicated network can improve the sample efficiency. In general, the capacity of the replay memory is limited and the replay memory is updated with the last transition of the agent. Thus, the agent mainly learns the recent transitions and gradually reaches the goal. However, as previously mentioned, using the bi-directional memory editing, we can obtain a number of sub-goals and behavior of the agent. Therefore, using the separated network and the replay memory, we can learn the agent to reach the sub-goals whenever, during or after the policy network learning.

Although the agent of the policy network cannot reach the final policy, the agent of the sub-goals dedicated network can reach the various sub-goals as well as the final goal. The users can fully control the agent by collecting various sub-goals from the bi-directional memory editing and by learning the agent on the sub-goals dedicated network.

\subsection{Reward shaping for the shorter path}
In the path planning, one of the important factors is to make the agent reach the destination within a specific period. To improve the agent’s ability, it is necessary to give enough rewards according to the steps to reach the target point. However, in this study, I focus on learning the sub-goals, and additional rewards are given to the agent to reach the sub-goals, regardless of the shortest path. Furthermore, I want to confirm that the agent can reach various sub-points in the environment. Thus, I assume that the environment of this study is a sparse reward environment so that the agent does not need to reach the target point in the shortest path. Rather, due to the exploration bonus, it is likely for the agent to delay the one’s arrival. Therefore, we propose the reward shaping when the bi-directional editing is performed for the shorter path of the agent in the path planning.
When the bi-directional editing is performed, the additional rewards ($r_t'$) are given with the corresponding sub-goals. Here, the rewards are reshaped as follows: 
\begin{eqnarray}
{r_{tg}} & = & r_{t}' + (dist_{short} – dist_{s_t})
\end{eqnarray}
: where $dist_{short}$ indicates the number of the shortest steps that can be possibly reached by the agent from $s_t$ to the current sub-goal ($g_s$), and $dist_{st}$ indicates the number of steps of the path that is moved by the agent from $s_t$ to the current sub-goal ($g_s$). That is, in each sub-goal, the agent gets a penalty by the number of steps to reach the sub-goal point. If the agent reaches the sub-goal in the shortest path, the agent does not receive any penalty. Otherwise, the agent receives a penalty depending on the number of steps to reach the sub-goal points.

\subsection{Learning a fully controllable agent in the path planning}
Fig \ref{fig2} shows the summary of the proposed method. The inverse module predicts the action of when $s_t$ and $s_{t+1}$ are given in the bi-directional memory editing. The sub-goals from two trajectories are generated. At this time, the reward shaping for the shorter path is performed. Then, the transitions are stored in the replay memory for the sub-goals ($\mathcal{D_s}$). The sub-goals dedicated network is trained independently with the policy network, whereas the policy network is only trained for the final policy. After learning is completed, in the various scenarios, the agent gets the sub-goals that are defined by the users and tried to achieve the sub-goals as well as the final goal. 

Algorithm 1 shows the procedure in the proposed method in detail. The simulation stage is similar to the other RL methods except for the inverse module to obtain $\hat{a}_{t+1}$ by predict $a_{t+1}’$ when $s_t$ and $s_{t+1}$ are given. In the bi-directional memory editing, the reverse transition $\{(s_{t+1},\hat{a}_{t+1},r_t,s_{t})\}$ is obtained using the inverse module and various sub-goals are generated and stored in the memory $(D_s)$. In the learning stage, optimize the policy network $(\theta_p)$ for the final goal and optimize the sub-goals dedicated network  $(\theta_g)$  for the sub-goals. In fact, it does not matter to learn the goals dedicated network separately after learning the policy network is completed, if the number of the edited memories is enough.

\section{Experiments}
\subsection{Experimental setting}
In the experiments, I wanted to confirm that the agent can be fully controlled by the sub-goals and that the agent can achieve various sub-goals, which were never visited by the agent in the learning. Thus, the training environment was constructed in a simpler way, while the test environment in a difficult way. I assumed a number of scenarios to test the agent’s ability of the path searching. Furthermore, I wanted to show the effect of the reward shaping for the shorter path.

  Fig \ref{fig1} shows the first environment. The goal of the agent was set to reach the target point in a simple two-dimensional (2D) grid environment. The reward was 0 except for the agent reaching the target point (+30). The RL model was performed for a total of 10,000 episodes. In the test environment, I set a total of 26 scenarios. In each scenario, several sub-goals were given to the agent. At first, a sub-goal nearest the current location of the agent was given to the agent. Then, if the agent reaches the sub-goal, the next sub-goal nearest the current location was given. Also, if the agent could not reach the given sub-goal, the next sub-goal nearest the current location was given.
 In various scenarios, I observed whether the agent reaches various sub-goals and successfully performs difficult missions such as a round trip. Moreover, I compared with and without the reward shaping for the shorter path in the proposed method. In each scenario, I calculated the number of steps to reach the target point.

\begin{figure}[t]
\centering
\includegraphics[scale=0.35]{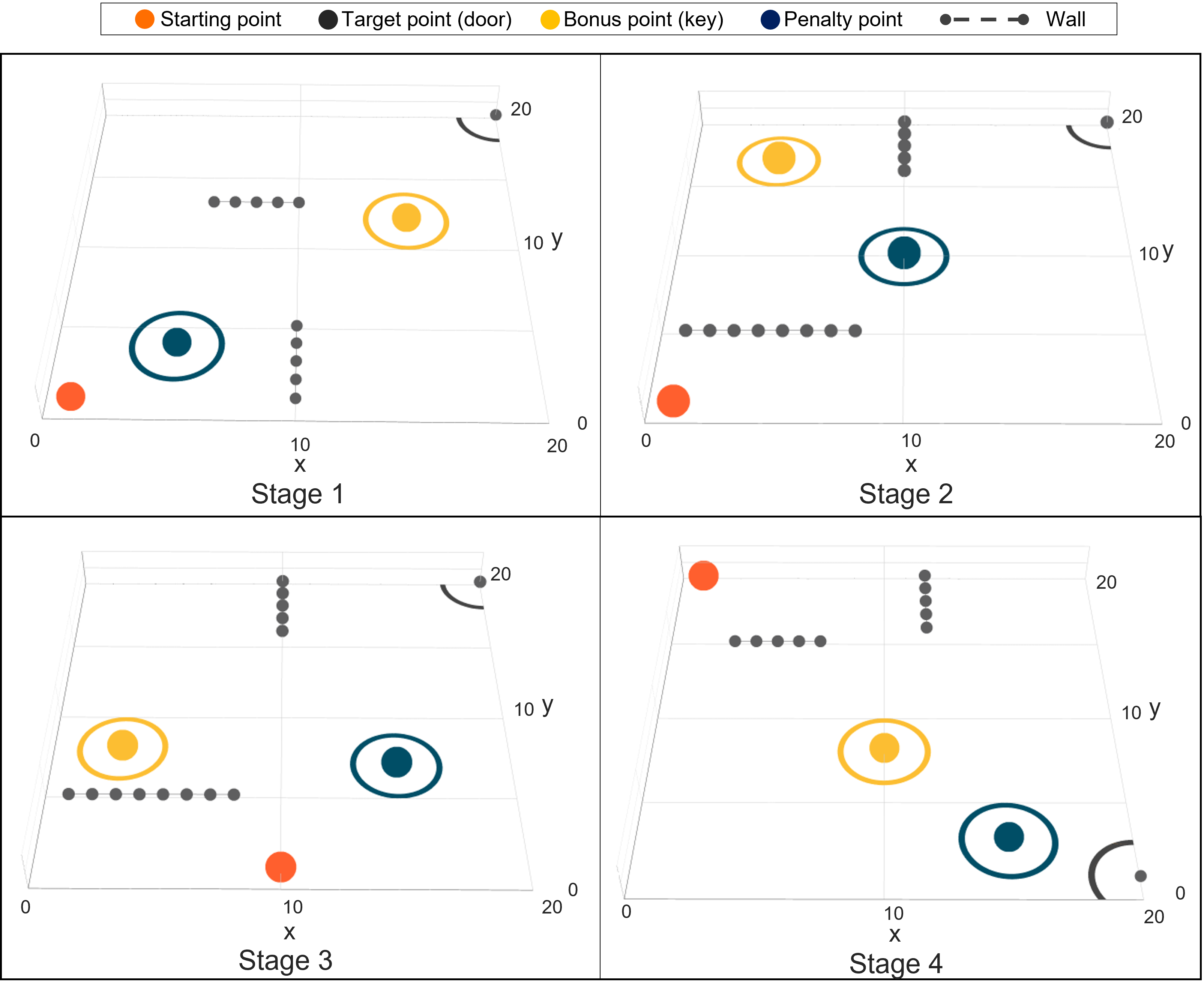}
\caption{The key-door domain environment. In each stage, the factors (a starting point, a bonus point, a wall, and a target point) were set differently. The agent must go through the bonus point to clear the stage. Even if the agent reaches the target point if the agent did not go through the bonus point, the agent cannot pass by the next stage.}
\label{fig3}
\end{figure}

  Fig \ref{fig3} shows the second environment. The environment is the ‘key-door domain’. The environment has a total of 4 stages and the agent should go through the bonus point (key) to clear the stage (door). Even though the agent reaches the target point (door), if the agent could not pass the bonus point (key), the agent cannot jump up to the next stage. The reward was set +10 for a bonus point, -10 for a penalty point, and +100 for the goal, when the agent goes through these points, respectively. This environment was difficult because of the condition that the agent must pass the bonus point to clear each stage and that the environment was defined as the sparse reward setting. The RL model performed for a total of 50,000 episodes. In the test environment, I set the two scenarios. In each stage, two sub-goals, defined by the user, the bonus point, and the goal, were given to the agent as the sub-goals in an order.

\subsection{Base architecture of the RL model}

Self-imitation learning (SIL) is an on-policy algorithm to exploit valuable past decisions from the replay memory \cite{oh2018self}. In the learning stage, the transitions are sampled, and they are trained by the policy network. At this time, if the transitions of the past are not valuable compared to the current value, the transitions are not exploited. That is, the SIL imitates valuable behaviors of the agent in the past. The authors combined the SIL and the on-policy RL model \cite{mnih2016asynchronous,schulman2017proximal} and proposed the following off-policy actor-critic loss:

\begin{eqnarray}
\mathcal{L}&=&\mathbb{E}_{s,a,R\in \mathcal{D}}[\mathcal{L}_{policy} + \beta\mathcal{L}_{value}] , \label{eq:sil1} \\
\mathcal{L}_{policy}&=&-log\pi_{\theta}(a|s)(R-V_\theta(S))_+  , \label{eq:sil2} \\
\mathcal{L}_{value}& =&{{1}\over{2}} \parallel(R-V_\theta(S))_+\parallel ^2 , \label{eq:sil3}
\end{eqnarray}
where $(\cdot)_+=max(\cdot,0)$; $\pi_\theta$ and $V_{\theta}(s)$ are the policy network and the value network, respectively, parameterized by $\theta$. The value loss is controlled by $\beta \in \mathbb{R}^+$. From the $(\cdot)_+$ operator in the loss, the transition, in which the current value is larger than the past return, is trained by the policy network and the value network.

The reason why that the SIL is used in the study is that the exploitation of valuable transition is needed. In fact, in the study, the off-policy RL model is necessary to utilize the replay memory \cite{mnih2015human,schaul2015prioritized,van2016deep}. However, in the path planning, to reach the target point, various routes can be obtained. Accordingly, I utilize the off-policy actor-critic RL model to get an effect of both the on-policy and the off-policy. The final RL architecture in this study is the combination of the SIL and the actor-critic network (ASIL).

In addition, I utilized the random network distillation (RND), which is widely used as an exploration bonus method \cite{burda2018exploration}. The RND uses two networks: a fixed and random initialized network (target network), and a predictor network trained using the output of the target network. The exploration bonus is given as a difference between the outputs of the two networks. If the agent visits a specific point continually, the exploration bonus is gradually decreased. Otherwise, if the agent visits a novel space, a large exploration bonus will be given to the agent.

\begin{figure}[t]
\centering
\includegraphics[scale=0.45]{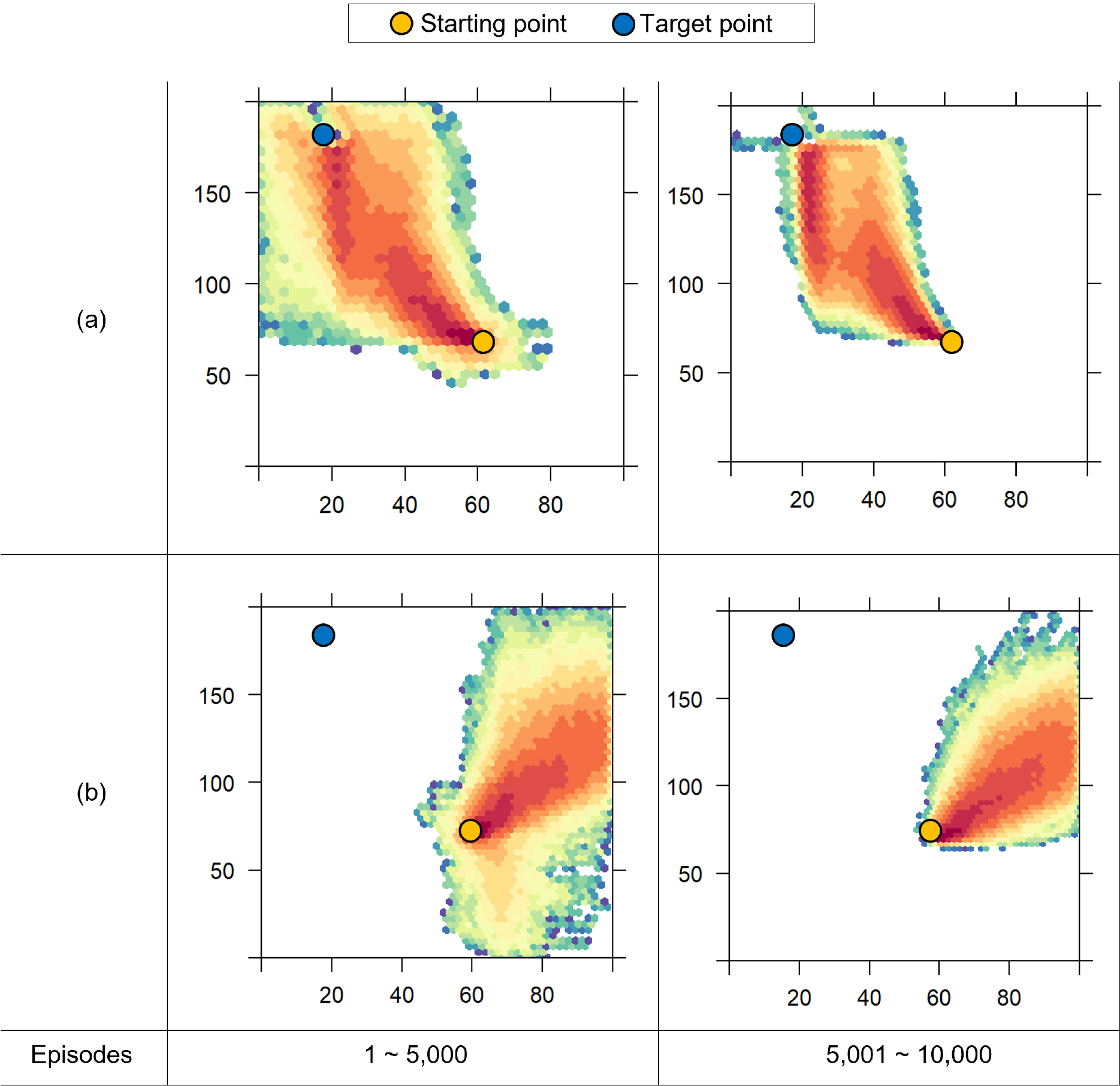}
\caption{The visualization of the path of the agent for the first environment. A color change from blue to red indicates that the agent has visited more often. \textbf{a}, The path of the agent of the policy network without learning the sub-goals. The agent easily reached the target point. \textbf{b}, The path of the agent of policy network with learning the sub-goals using the bi-directional memory editing. The agent was confused by the sub-goals, so that the agent could not reach the final goal. }
\label{fig4}
\end{figure}

\subsection{Simple 2D grid environment}
Fig \ref{fig4} shows the visualization of the path that is moved by the agent of the policy network without (a) and with (b) learning the sub-goals. A color change from blue to red means that the agent has visited more often. When the policy network did not learn the sub-goals, the agent of the policy network easily reached the target point, and the agent mostly moved only between the starting point and the target point. That is, the agent mostly moved the left and top areas in the environment because of the location of the target point.

\begin{figure*}[h]
\centering
\includegraphics[scale=0.45]{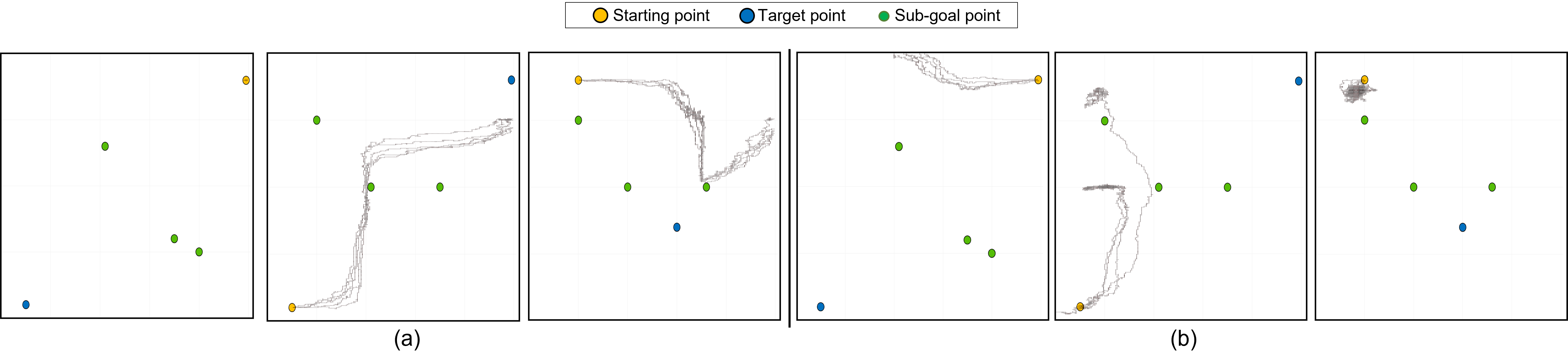}
\caption{\textbf{a}, The visualization of the path of the agent of the policy network with learning the sub-goals. The agent could not learn the sub-goals so the agent had a tendency to just move to the right. \textbf{b},  The visualization of the path of the agent of the sub-dedicated network with the forward directional memory editing. The agent was trained using a one-directional route. Therefore, the agent tried leftward, even though the given sub-goals were located on the right of the agent. }
\label{fig6}
\end{figure*}

\begin{figure*}[h]
\centering
\includegraphics[scale=0.42]{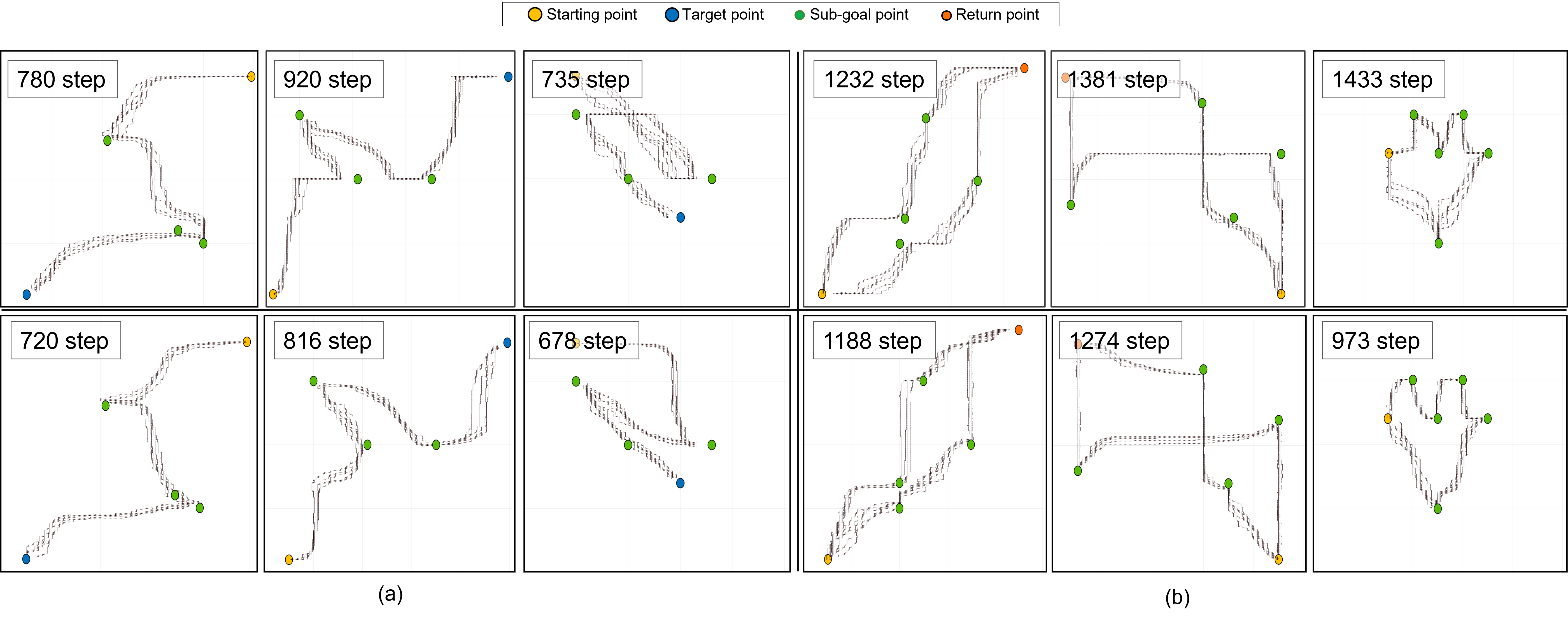}
\caption{The visualization of the path of the agent when the sub-goals were given in the test environment according to the reward shaping. The top and the bottom of the figure indicate the route of the agent without and with the reward shaping, respectively. (a), The agent successfully went through the sub-goals and reached the final goal. (b), In a round-trip environment, the agent also passed all the sub-goals and came back to the starting point. With the reward shaping (bottom), the agent arrived at each sub-goal point and the final goal faster than without the reward shaping (top). }
\label{fig5}
\end{figure*}

In addition, if the policy network was trained on the sub-goals using the bi-directional memory editing, the agent of the policy network could not reach the target point, as shown in Fig \ref{fig4}.b. At a specific point, if the agent is trained to go in various directions by the sub-goals, the agent of the policy network was confused about where it has to go. Then, the agent would fail to reach the final goal. In the experiment, the agent just moved to the right area in the environment. This case was repeated every time the network was trained. Notably, in the experiment, the agent of the policy network without learning the sub-goals always could reach the final goal. In the test environment, the agent of the policy network, which learns the sub-goals using  bi-directional editing, also failed to reach the sub-goals, as shown in Fig \ref{fig6}.a. The agent even left the environment as soon as the agent departed, as shown in Fig \ref{fig6}.a.(left). This is because the agent of the policy network mainly moved to the right area in the environment due to the confusion.

When the sub-goals dedicated network was trained using the forward directional memory editing only, the agent failed to reach the sub-goals, as shown in Fig \ref{fig6}.b. It was observed that the agent tried to move in the left direction. The reason is that the agent of the policy network mainly moved to the leftward and the upward direction, and the agent is trained using the one-directional trajectories. 

Fig \ref{fig5} shows the result of learning the sub-goals in the test environments without (top) and with (bottom) the reward shaping for the shorter path. The agent was trained from the sub-goals dedicated network using the trajectories, which were collected from the agent of the policy network. In the test environment, I set the sub-goals difficultly, even though all the points including the starting point and the target point were set inversely. The agent successfully reached all the sub-goals and the target points in the experiment, as shown in Fig \ref{fig5}.a. That is, the agent was able to be fully controlled by the sub-goals, in various scenarios. The agent reached the sub-goals that had never been visited by the agent of the policy network. Interestingly, in extremely hard environments (round trip tasks), as shown in Fig \ref{fig5}.b, the agent departed from the starting point and went through the sub-goals, and then the agent turned halfway point and came back to the starting point. 

\begin{figure*}[t]
\centering
\includegraphics[scale=0.55]{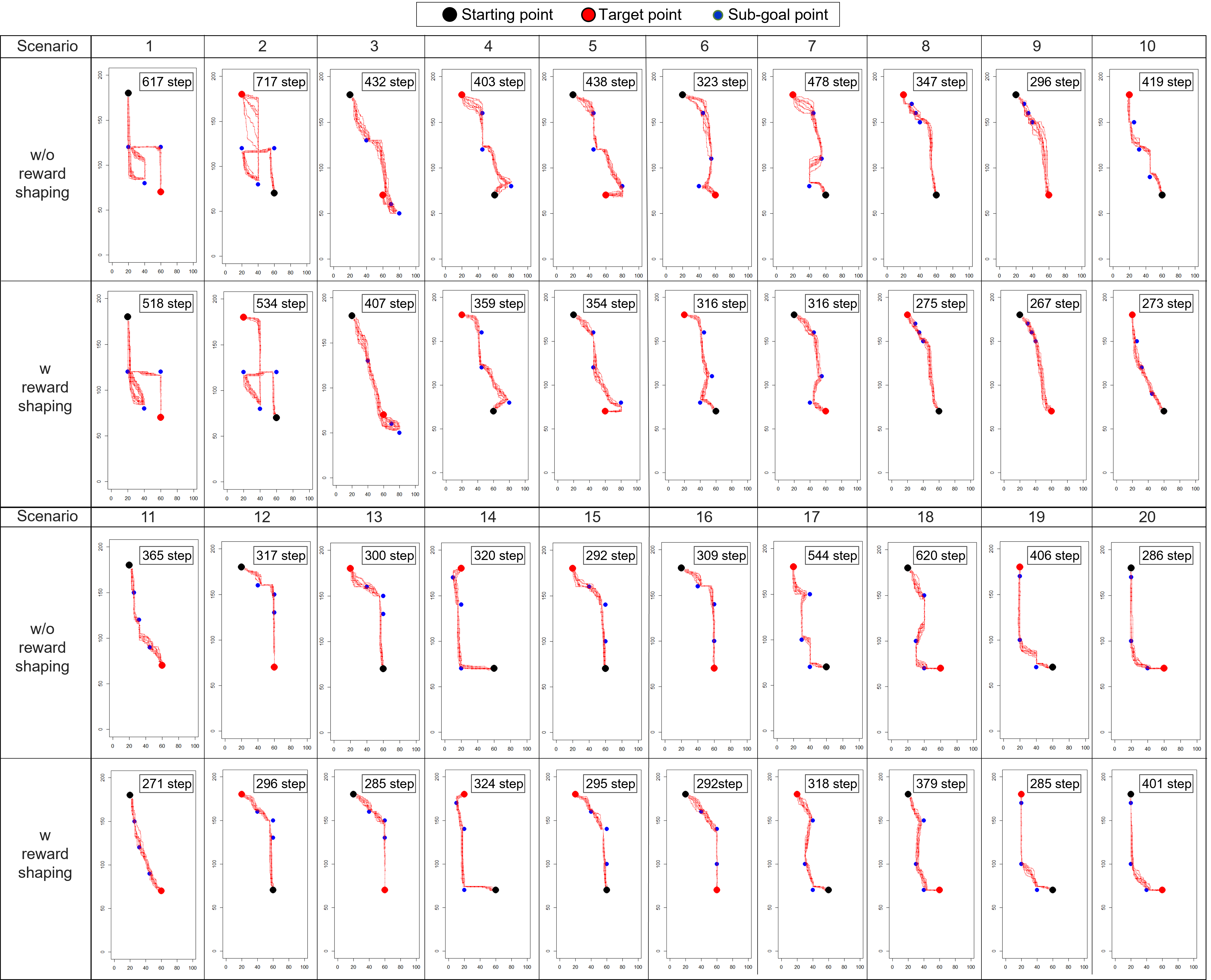}
\caption{The visualization of the path of the agent without and with the reward shaping for the shorter path in 20 scenarios. The number, located in the upper of each figure, indicates the number of steps to reach the target point. The agent of the sub-goals dedicated network arrived at the target point in all scenarios. With a visual inspection, the path of the agent with the reward shaping was shorter than without the reward shaping. In the comparison of the number of steps, it can be seen that the agent with the reward shaping reaches the target point within a short time than the agent without the reward shaping. }
\label{fig7}
\end{figure*}

In addition, with the reward shaping for the shorter path, the agent reached the target point faster than without the reward shaping, as shown in Fig \ref{fig5}. With a visual inspection, we can observe that the agent with the reward shaping reached the sub-goals located diagonally with a shorter distance from the current location. Moreover, the difference between the number of steps to reach the target point with and without the reward shaping was significant.

To strongly confirm the performance difference between without and with the reward shaping of the proposed method, I additionally constructed 20 scenarios. Fig \ref{fig7} shows the route of the agent with and without the reward shaping in each scenario. The number in each figure indicates the number of steps to reach the target point. In all scenarios, the agent went through the sub-goals and reached the final goal. It can be seen that the agent was able to be fully under control by the sub-goals. Moreover, except for three scenarios, the scenario 14, 15, and 16, the agent with the reward shaping reached the target point much faster in all scenarios. This characteristic was salient in complex environments such as the scenario 1 and 2. The average of the steps with the reward shaping was 338.25 and the average of the steps without the reward shaping was 411.45. It was confirmed that the rewards shaping can shorten the number of steps to arrive at the destination by 21.6.

\subsection{Key-door domain}
The agent of the (ASIL + RND) reached the final stage within 50,000 episodes only one time out of 10 trials in the ‘key-door domain’ environment. It was a very difficult environment to clear. This was because of the condition to clear each stage and the sparseness of the reward. I assumed the two scenarios in the test environment. Two sub-goals were imposed on the agent differently in each stage. The bonus point and the goal point are also given as the sub-goals. The agent was enforced to reach the sub-goals at first and after that, the agent was encouraged to reach the bonus point and the target point. 

\begin{figure*}[h]
\centering
\includegraphics[width=\textwidth]{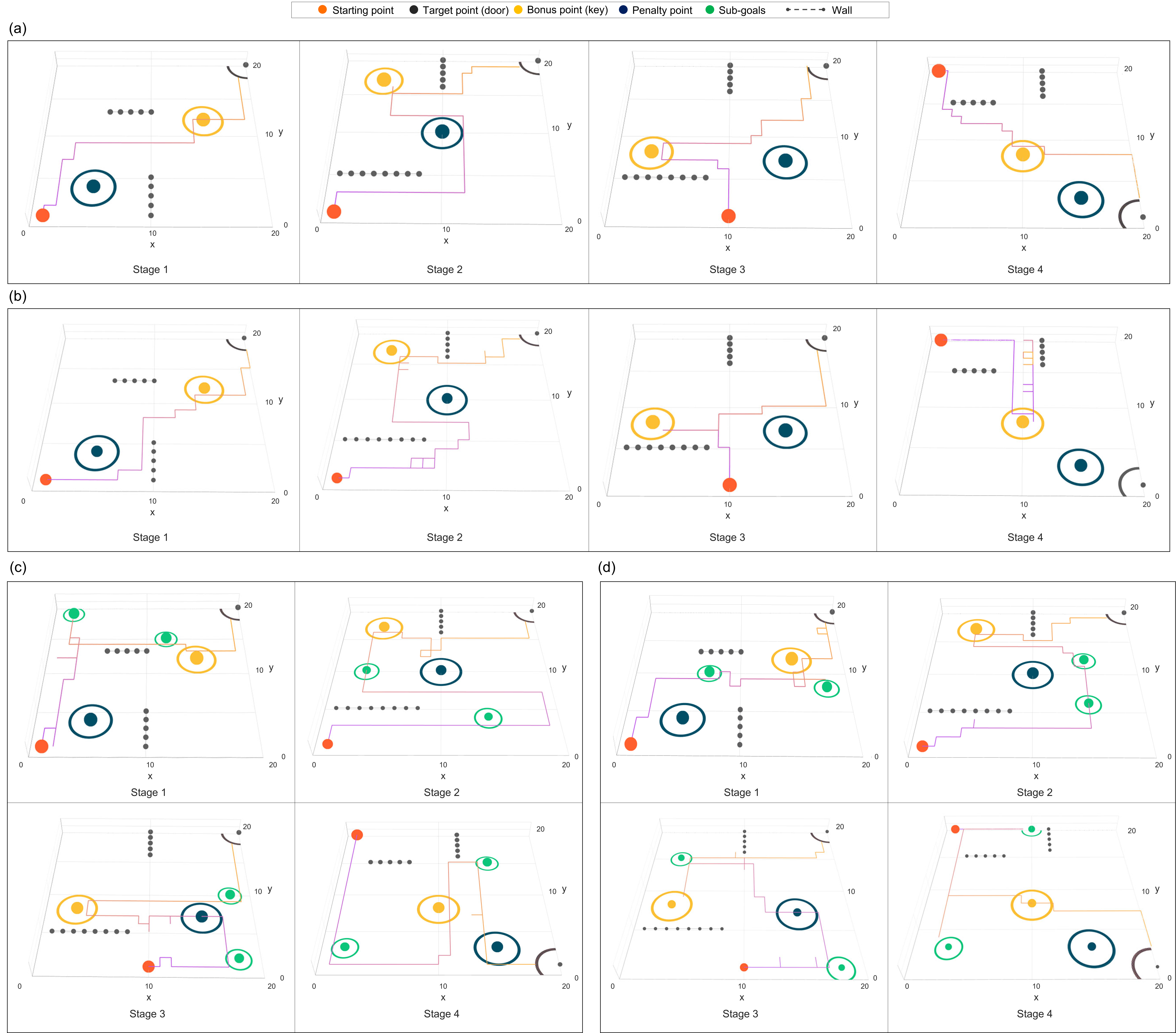}
\caption{The visualization of the path of the agent in the key-door domain environment. \textbf{a}, The path of the agent of the policy network in each stage, when the agent cleared all stages. The agent of the policy network cleared only one time out of 10 trials in total. \textbf{b}, The path of the agent of the policy network in each stage, when the agent did not clear all stages. \textbf{c} $\sim$ \textbf{d}, The path of the agent of the sub-goals dedicated network, when the sub-goals were given. The sub-goals dedicated network was used, when the agent of the policy network failed to clear all stages.}
\label{fig8}
\end{figure*}

Fig \ref{fig8}.a shows the visualization of the path of the agent, success case of learning the desired policy. If the policy network converges to the desired policy, the agent of the network showed a clear path to the goal of Stage 4. However, in the experiment, the agent of the policy network almost failed to reach the final stage in the training environment within 50,000 episodes. Fig \ref{fig8}.b shows the visualization of the path of the agent, when the agent of the policy network failed to learn the desired policy. The agent could reach Stage 4, but the agent failed to clear the stage. In contrast, the agent of the sub-goals dedicated network was able to reach the goal in the final stage, going through the sub-goals and the bonus point in the two scenarios as shown in Fig \ref{fig8}.c $\sim$ d, even though the agent of the policy network failed, as shown in Fig \ref{fig8}.b. This result means that learning the sub-goals can improve the ability of the agent to reach the final goal, just like in previous studies. Unlike the previous study, I set the mission for the agent’s performance to the sub-goals dedicated network in order to apply  in the path planning problem, in which the ability to move in various directions is necessary. Indeed, in the experiments, the last 300 episodes of the policy network were enough to learn the sub-goals dedicated network so that the agent could be under control. We do not need to collect a number of trajectories to learn the agent in the path planning, if we utilize the proposed method.

However, the agent of the sub-goals dedicated network did not show the shortest path whereas the agent of the policy network showed the almost shortest path. Furthermore, like a previous study \cite{lee2022learning}, sometimes, the agent that learned the sub-goals was confused when the bonus point was near the current location of the agent. The reason is that the agent was trained to go through the bonus point, at first, to clear each stage. That is, the agent was enforced to move to the sub-goals in learning the sub-goals, and the agent was also encouraged to move to the bonus point and the target point. When the agent was near the bonus point, the agent could get a larger reward, although the agent did not reach the sub-goals. The agent was partially able to be controllable in a complex environment with various variables that have a great deal of influence on the agent. These remaining issues call for further studies.

\section{Conclusion}
In this paper, I propose a novel RL framework within which the agent can be under control in the path planning environment so that the agent can reach various sub-goals. The agent that completed the learning can perform the difficult missions such as a round trip, even the agent can reach an unknown area. Therefore, the bi-directional memory editing and the sub-goals dedicated network were presented on the goal-conditioned RL. From the bi-directional memory editing, we can obtain various sub-goals and behaviors of the agent such that the agent can be more robust in the test environment. In addition, using the sub-goals dedicated network, the agent can perform several behaviors that are directed by different sub-goals at one point. It was confirmed that the agent can be fully controlled and can achieve various sub-goals that are customized by the users in the test environment. Furthermore, the proposed reward shaping for the shorter path can improve the ability of the path planning. 

However, in a complex environment with various variables such as the key-door domain, the agent was confused about whether to select the sub-goals or the bonus point. Although a fully controllable agent is useful in the path planning, various variables such as an obstacle and the limited number of steps in an environmental setting should be considered. Further, the reward shaping cannot guarantee the optimal path. Future studies are required for a fully controllable agent for the path planning. I expect that this study would be applied and studied in a variety of domains with a fully controllable agent in various scenarios.

\ifCLASSOPTIONcompsoc
  \section*{Acknowledgments}
\else
  \section*{Acknowledgment}
\fi


\ifCLASSOPTIONcaptionsoff
  \newpage
\fi



%

\bibliographystyle{IEEEtran}

\bibliography{mybibfile}

\begin{thebibliography}{10}
\providecommand{\url}[1]{#1}
\csname url@samestyle\endcsname
\providecommand{\newblock}{\relax}
\providecommand{\bibinfo}[2]{#2}
\providecommand{\BIBentrySTDinterwordspacing}{\spaceskip=0pt\relax}
\providecommand{\BIBentryALTinterwordstretchfactor}{4}
\providecommand{\BIBentryALTinterwordspacing}{\spaceskip=\fontdimen2\font plus
\BIBentryALTinterwordstretchfactor\fontdimen3\font minus
  \fontdimen4\font\relax}
\providecommand{\BIBforeignlanguage}[2]{{%
\expandafter\ifx\csname l@#1\endcsname\relax
\typeout{** WARNING: IEEEtran.bst: No hyphenation pattern has been}%
\typeout{** loaded for the language `#1'. Using the pattern for}%
\typeout{** the default language instead.}%
\else
\language=\csname l@#1\endcsname
\fi
#2}}
\providecommand{\BIBdecl}{\relax}
\BIBdecl

\bibitem{yu2020path}
J.~Yu, Y.~Su, and Y.~Liao, ``The path planning of mobile robot by neural
  networks and hierarchical reinforcement learning,'' \emph{Frontiers in
  Neurorobotics}, p.~63, 2020.

\bibitem{kumar2020comparison}
R.~Kumar, L.~Singh, and R.~Tiwari, ``Comparison of two meta--heuristic
  algorithms for path planning in robotics,'' in \emph{2020 International
  Conference on Contemporary Computing and Applications (IC3A)}.\hskip 1em plus
  0.5em minus 0.4em\relax IEEE, 2020, pp. 159--162.

\bibitem{wohlke2021hierarchies}
J.~W{\"o}hlke, F.~Schmitt, and H.~van Hoof, ``Hierarchies of planning and
  reinforcement learning for robot navigation,'' in \emph{2021 IEEE
  International Conference on Robotics and Automation (ICRA)}.\hskip 1em plus
  0.5em minus 0.4em\relax IEEE, 2021, pp. 10\,682--10\,688.

\bibitem{jeauneau2018path}
V.~Jeauneau, L.~Jouanneau, and A.~Kotenkoff, ``Path planner methods for uavs in
  real environment,'' \emph{IFAC-PapersOnLine}, vol.~51, no.~22, pp. 292--297,
  2018.

\bibitem{qie2019joint}
H.~Qie, D.~Shi, T.~Shen, X.~Xu, Y.~Li, and L.~Wang, ``Joint optimization of
  multi-uav target assignment and path planning based on multi-agent
  reinforcement learning,'' \emph{IEEE access}, vol.~7, pp. 146\,264--146\,272,
  2019.

\bibitem{wang2020multi}
L.~Wang, K.~Wang, C.~Pan, W.~Xu, N.~Aslam, and L.~Hanzo, ``Multi-agent deep
  reinforcement learning-based trajectory planning for multi-uav assisted
  mobile edge computing,'' \emph{IEEE Transactions on Cognitive Communications
  and Networking}, vol.~7, no.~1, pp. 73--84, 2020.

\bibitem{wang2020mobile}
B.~Wang, Z.~Liu, Q.~Li, and A.~Prorok, ``Mobile robot path planning in dynamic
  environments through globally guided reinforcement learning,'' \emph{IEEE
  Robotics and Automation Letters}, vol.~5, no.~4, pp. 6932--6939, 2020.

\bibitem{hayat2020multi}
S.~Hayat, E.~Yanmaz, C.~Bettstetter, and T.~X. Brown, ``Multi-objective drone
  path planning for search and rescue with quality-of-service requirements,''
  \emph{Autonomous Robots}, vol.~44, no.~7, pp. 1183--1198, 2020.

\bibitem{bayerlein2021multi}
H.~Bayerlein, M.~Theile, M.~Caccamo, and D.~Gesbert, ``Multi-uav path planning
  for wireless data harvesting with deep reinforcement learning,'' \emph{IEEE
  Open Journal of the Communications Society}, vol.~2, pp. 1171--1187, 2021.

\bibitem{lee2020autonomous}
G.~T. Lee and C.~O. Kim, ``Autonomous control of combat unmanned aerial
  vehicles to evade surface-to-air missiles using deep reinforcement
  learning,'' \emph{IEEE Access}, vol.~8, pp. 226\,724--226\,736, 2020.

\bibitem{hu2022autonomous}
J.~Hu, L.~Wang, T.~Hu, C.~Guo, and Y.~Wang, ``Autonomous maneuver decision
  making of dual-uav cooperative air combat based on deep reinforcement
  learning,'' \emph{Electronics}, vol.~11, no.~3, p. 467, 2022.

\bibitem{wang2019obstacle}
P.~Wang, S.~Gao, L.~Li, B.~Sun, and S.~Cheng, ``Obstacle avoidance path
  planning design for autonomous driving vehicles based on an improved
  artificial potential field algorithm,'' \emph{Energies}, vol.~12, no.~12, p.
  2342, 2019.

\bibitem{tammvee2021human}
M.~Tammvee and G.~Anbarjafari, ``Human activity recognition-based path planning
  for autonomous vehicles,'' \emph{Signal, Image and Video Processing},
  vol.~15, no.~4, pp. 809--816, 2021.

\bibitem{yao2020path}
Q.~Yao, Z.~Zheng, L.~Qi, H.~Yuan, X.~Guo, M.~Zhao, Z.~Liu, and T.~Yang, ``Path
  planning method with improved artificial potential field—a reinforcement
  learning perspective,'' \emph{IEEE Access}, vol.~8, pp. 135\,513--135\,523,
  2020.

\bibitem{qiao2020hierarchical}
Z.~Qiao, Z.~Tyree, P.~Mudalige, J.~Schneider, and J.~M. Dolan, ``Hierarchical
  reinforcement learning method for autonomous vehicle behavior planning,'' in
  \emph{2020 IEEE/RSJ International Conference on Intelligent Robots and
  Systems (IROS)}.\hskip 1em plus 0.5em minus 0.4em\relax IEEE, 2020, pp.
  6084--6089.

\bibitem{lin2021collision}
G.~Lin, L.~Zhu, J.~Li, X.~Zou, and Y.~Tang, ``Collision-free path planning for
  a guava-harvesting robot based on recurrent deep reinforcement learning,''
  \emph{Computers and Electronics in Agriculture}, vol. 188, p. 106350, 2021.

\bibitem{cao2021confidence}
Z.~Cao, S.~Xu, H.~Peng, D.~Yang, and R.~Zidek, ``Confidence-aware reinforcement
  learning for self-driving cars,'' \emph{IEEE Transactions on Intelligent
  Transportation Systems}, 2021.

\bibitem{li2019deep}
H.~Li, Q.~Zhang, and D.~Zhao, ``Deep reinforcement learning-based automatic
  exploration for navigation in unknown environment,'' \emph{IEEE transactions
  on neural networks and learning systems}, vol.~31, no.~6, pp. 2064--2076,
  2019.

\bibitem{yang2020efficient}
Y.~Yang, M.~A. Bevan, and B.~Li, ``Efficient navigation of colloidal robots in
  an unknown environment via deep reinforcement learning,'' \emph{Advanced
  Intelligent Systems}, vol.~2, no.~1, p. 1900106, 2020.

\bibitem{hu2020voronoi}
J.~Hu, H.~Niu, J.~Carrasco, B.~Lennox, and F.~Arvin, ``Voronoi-based
  multi-robot autonomous exploration in unknown environments via deep
  reinforcement learning,'' \emph{IEEE Transactions on Vehicular Technology},
  vol.~69, no.~12, pp. 14\,413--14\,423, 2020.

\bibitem{lee2022learning}
G.~Lee, ``Learning user-defined sub-goals using memory editing in reinforcement
  learning,'' \emph{arXiv preprint arXiv:2205.00399}, 2022.

\bibitem{lee2020weakly}
L.~Lee, B.~Eysenbach, R.~R. Salakhutdinov, S.~S. Gu, and C.~Finn,
  ``Weakly-supervised reinforcement learning for controllable behavior,''
  \emph{Advances in Neural Information Processing Systems}, vol.~33, pp.
  2661--2673, 2020.

\bibitem{okudo2021subgoal}
T.~Okudo and S.~Yamada, ``Subgoal-based reward shaping to improve efficiency in
  reinforcement learning,'' \emph{IEEE Access}, vol.~9, pp. 97\,557--97\,568,
  2021.

\bibitem{zhang2021world}
L.~Zhang, G.~Yang, and B.~C. Stadie, ``World model as a graph: Learning latent
  landmarks for planning,'' in \emph{International Conference on Machine
  Learning}.\hskip 1em plus 0.5em minus 0.4em\relax PMLR, 2021, pp.
  12\,611--12\,620.

\bibitem{chane2021goal}
E.~Chane-Sane, C.~Schmid, and I.~Laptev, ``Goal-conditioned reinforcement
  learning with imagined subgoals,'' in \emph{International Conference on
  Machine Learning}.\hskip 1em plus 0.5em minus 0.4em\relax PMLR, 2021, pp.
  1430--1440.

\bibitem{yang2022rethinking}
R.~Yang, Y.~Lu, W.~Li, H.~Sun, M.~Fang, Y.~Du, X.~Li, L.~Han, and C.~Zhang,
  ``Rethinking goal-conditioned supervised learning and its connection to
  offline rl,'' \emph{arXiv preprint arXiv:2202.04478}, 2022.

\bibitem{bai2019guided}
C.~Bai, P.~Liu, W.~Zhao, and X.~Tang, ``Guided goal generation for hindsight
  multi-goal reinforcement learning,'' \emph{Neurocomputing}, vol. 359, pp.
  353--367, 2019.

\bibitem{zhao2019maximum}
R.~Zhao, X.~Sun, and V.~Tresp, ``Maximum entropy-regularized multi-goal
  reinforcement learning,'' in \emph{International Conference on Machine
  Learning}.\hskip 1em plus 0.5em minus 0.4em\relax PMLR, 2019, pp. 7553--7562.

\bibitem{bernstein2009tell}
D.~M. Bernstein and E.~F. Loftus, ``How to tell if a particular memory is true
  or false,'' \emph{Perspectives on Psychological Science}, vol.~4, no.~4, pp.
  370--374, 2009.

\bibitem{schacter2011memory}
D.~L. Schacter, S.~A. Guerin, and P.~L.~S. Jacques, ``Memory distortion: An
  adaptive perspective,'' \emph{Trends in cognitive sciences}, vol.~15, no.~10,
  pp. 467--474, 2011.

\bibitem{fernandez2015benefits}
J.~Fern{\'a}ndez, ``What are the benefits of memory distortion?'' 2015.

\bibitem{phelps2019memory}
E.~A. Phelps and S.~G. Hofmann, ``Memory editing from science fiction to
  clinical practice,'' \emph{Nature}, vol. 572, no. 7767, pp. 43--50, 2019.

\bibitem{yan2018path}
C.~Yan and X.~Xiang, ``A path planning algorithm for uav based on improved
  q-learning,'' in \emph{2018 2nd International Conference on Robotics and
  Automation Sciences (ICRAS)}.\hskip 1em plus 0.5em minus 0.4em\relax IEEE,
  2018, pp. 1--5.

\bibitem{chen2020improved}
J.~Chen, M.~Li, Z.~Yuan, and Q.~Gu, ``An improved a* algorithm for uav path
  planning problems,'' in \emph{2020 IEEE 4th Information Technology,
  Networking, Electronic and Automation Control Conference (ITNEC)},
  vol.~1.\hskip 1em plus 0.5em minus 0.4em\relax IEEE, 2020, pp. 958--962.

\bibitem{zhou2020trajectory}
H.~Zhou, H.-L. Xiong, Y.~Liu, N.-D. Tan, and L.~Chen, ``Trajectory planning
  algorithm of uav based on system positioning accuracy constraints,''
  \emph{Electronics}, vol.~9, no.~2, p. 250, 2020.

\bibitem{huang2018uav}
C.~Huang and J.~Fei, ``Uav path planning based on particle swarm optimization
  with global best path competition,'' \emph{International Journal of Pattern
  Recognition and Artificial Intelligence}, vol.~32, no.~06, p. 1859008, 2018.

\bibitem{chen2021three}
J.~Chen, H.~Zhao, and L.~Wang, ``Three dimensional path planning of uav based
  on adaptive particle swarm optimization algorithm,'' in \emph{Journal of
  Physics: Conference Series}, vol. 1846, no.~1.\hskip 1em plus 0.5em minus
  0.4em\relax IOP Publishing, 2021, p. 012007.

\bibitem{wang2022improved}
X.~Wang, C.~Huang, and F.~Chen, ``An improved particle swarm optimization
  algorithm for unmanned aerial vehicle route planning,'' in \emph{Journal of
  Physics: Conference Series}, vol. 2245, no.~1.\hskip 1em plus 0.5em minus
  0.4em\relax IOP Publishing, 2022, p. 012013.

\bibitem{jamshidi2020analysis}
V.~Jamshidi, V.~Nekoukar, and M.~H. Refan, ``Analysis of parallel genetic
  algorithm and parallel particle swarm optimization algorithm uav path
  planning on controller area network,'' \emph{Journal of Control, Automation
  and Electrical Systems}, vol.~31, no.~1, pp. 129--140, 2020.

\bibitem{lei2018dynamic}
X.~Lei, Z.~Zhang, and P.~Dong, ``Dynamic path planning of unknown environment
  based on deep reinforcement learning,'' \emph{Journal of Robotics}, vol.
  2018, 2018.

\bibitem{liu2021novel}
X.~Liu, D.~Zhang, T.~Zhang, Y.~Cui, L.~Chen, and S.~Liu, ``Novel best path
  selection approach based on hybrid improved a* algorithm and reinforcement
  learning,'' \emph{Applied Intelligence}, vol.~51, no.~12, pp. 9015--9029,
  2021.

\bibitem{yan2020towards}
C.~Yan, X.~Xiang, and C.~Wang, ``Towards real-time path planning through deep
  reinforcement learning for a uav in dynamic environments,'' \emph{Journal of
  Intelligent \& Robotic Systems}, vol.~98, no.~2, pp. 297--309, 2020.

\bibitem{chen2019knowledge}
C.~Chen, X.-Q. Chen, F.~Ma, X.-J. Zeng, and J.~Wang, ``A knowledge-free path
  planning approach for smart ships based on reinforcement learning,''
  \emph{Ocean Engineering}, vol. 189, p. 106299, 2019.

\bibitem{guo2020autonomous}
S.~Guo, X.~Zhang, Y.~Zheng, and Y.~Du, ``An autonomous path planning model for
  unmanned ships based on deep reinforcement learning,'' \emph{Sensors},
  vol.~20, no.~2, p. 426, 2020.

\bibitem{andrychowicz2017hindsight}
M.~Andrychowicz, F.~Wolski, A.~Ray, J.~Schneider, R.~Fong, P.~Welinder,
  B.~McGrew, J.~Tobin, O.~Pieter~Abbeel, and W.~Zaremba, ``Hindsight experience
  replay,'' \emph{Advances in neural information processing systems}, vol.~30,
  2017.

\bibitem{nguyen2019hindsight}
H.~Nguyen, H.~M. La, and M.~Deans, ``Hindsight experience replay with
  experience ranking,'' in \emph{2019 Joint IEEE 9th International Conference
  on Development and Learning and Epigenetic Robotics (ICDL-EpiRob)}.\hskip 1em
  plus 0.5em minus 0.4em\relax IEEE, 2019, pp. 1--6.

\bibitem{fang2019curriculum}
M.~Fang, T.~Zhou, Y.~Du, L.~Han, and Z.~Zhang, ``Curriculum-guided hindsight
  experience replay,'' \emph{Advances in neural information processing
  systems}, vol.~32, 2019.

\bibitem{lai2020hindsight}
Y.~Lai, W.~Wang, Y.~Yang, J.~Zhu, and M.~Kuang, ``Hindsight planner,'' in
  \emph{Proceedings of the 19th International Conference on Autonomous Agents
  and MultiAgent Systems}, 2020, pp. 690--698.

\bibitem{nasiriany2019planning}
S.~Nasiriany, V.~Pong, S.~Lin, and S.~Levine, ``Planning with goal-conditioned
  policies,'' \emph{Advances in Neural Information Processing Systems},
  vol.~32, 2019.

\bibitem{ghosh2019learning}
D.~Ghosh, A.~Gupta, A.~Reddy, J.~Fu, C.~Devin, B.~Eysenbach, and S.~Levine,
  ``Learning to reach goals via iterated supervised learning,'' \emph{arXiv
  preprint arXiv:1912.06088}, 2019.

\bibitem{nair2018visual}
A.~V. Nair, V.~Pong, M.~Dalal, S.~Bahl, S.~Lin, and S.~Levine, ``Visual
  reinforcement learning with imagined goals,'' \emph{Advances in neural
  information processing systems}, vol.~31, 2018.

\bibitem{eysenbach2019search}
B.~Eysenbach, R.~R. Salakhutdinov, and S.~Levine, ``Search on the replay
  buffer: Bridging planning and reinforcement learning,'' \emph{Advances in
  Neural Information Processing Systems}, vol.~32, 2019.

\bibitem{eysenbach2020c}
B.~Eysenbach, R.~Salakhutdinov, and S.~Levine, ``C-learning: Learning to
  achieve goals via recursive classification,'' \emph{arXiv preprint
  arXiv:2011.08909}, 2020.

\bibitem{kim2021landmark}
J.~Kim, Y.~Seo, and J.~Shin, ``Landmark-guided subgoal generation in
  hierarchical reinforcement learning,'' \emph{Advances in Neural Information
  Processing Systems}, vol.~34, 2021.

\bibitem{nachum2018data}
O.~Nachum, S.~S. Gu, H.~Lee, and S.~Levine, ``Data-efficient hierarchical
  reinforcement learning,'' \emph{Advances in neural information processing
  systems}, vol.~31, 2018.

\bibitem{gurtler2021hierarchical}
N.~G{\"u}rtler, D.~B{\"u}chler, and G.~Martius, ``Hierarchical reinforcement
  learning with timed subgoals,'' \emph{Advances in Neural Information
  Processing Systems}, vol.~34, 2021.

\bibitem{oh2018self}
J.~Oh, Y.~Guo, S.~Singh, and H.~Lee, ``Self-imitation learning,'' in
  \emph{International Conference on Machine Learning}.\hskip 1em plus 0.5em
  minus 0.4em\relax PMLR, 2018, pp. 3878--3887.

\bibitem{mnih2016asynchronous}
V.~Mnih, A.~P. Badia, M.~Mirza, A.~Graves, T.~Lillicrap, T.~Harley, D.~Silver,
  and K.~Kavukcuoglu, ``Asynchronous methods for deep reinforcement learning,''
  in \emph{International conference on machine learning}.\hskip 1em plus 0.5em
  minus 0.4em\relax PMLR, 2016, pp. 1928--1937.

\bibitem{schulman2017proximal}
J.~Schulman, F.~Wolski, P.~Dhariwal, A.~Radford, and O.~Klimov, ``Proximal
  policy optimization algorithms,'' \emph{arXiv preprint arXiv:1707.06347},
  2017.

\bibitem{mnih2015human}
V.~Mnih, K.~Kavukcuoglu, D.~Silver, A.~A. Rusu, J.~Veness, M.~G. Bellemare,
  A.~Graves, M.~Riedmiller, A.~K. Fidjeland, G.~Ostrovski \emph{et~al.},
  ``Human-level control through deep reinforcement learning,'' \emph{nature},
  vol. 518, no. 7540, pp. 529--533, 2015.

\bibitem{schaul2015prioritized}
T.~Schaul, J.~Quan, I.~Antonoglou, and D.~Silver, ``Prioritized experience
  replay,'' \emph{arXiv preprint arXiv:1511.05952}, 2015.

\bibitem{van2016deep}
H.~Van~Hasselt, A.~Guez, and D.~Silver, ``Deep reinforcement learning with
  double q-learning,'' in \emph{Proceedings of the AAAI conference on
  artificial intelligence}, vol.~30, no.~1, 2016.

\bibitem{burda2018exploration}
Y.~Burda, H.~Edwards, A.~Storkey, and O.~Klimov, ``Exploration by random
  network distillation,'' \emph{arXiv preprint arXiv:1810.12894}, 2018.

\end{thebibliography}

%

\begin{IEEEbiography}[{\includegraphics[width=1in,height=1.25in,clip,keepaspectratio]{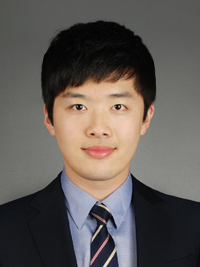}}]{GyeongTaek Lee}
received a Ph.D degree in industrial engineering from Yonsei University, South Korea, in 2022. He is a visiting researcher in Department of Industrial and Systems Engineering from Rutgers University, The State University of New Jersey, NJ, USA.  His current research interests include reinforcement learning and manufacturing data science. 
\end{IEEEbiography}






\end{document}